\documentclass{svproc}

\usepackage{url}
\usepackage{units}
\usepackage{graphicx}
\usepackage{cite}
\usepackage{amsmath}
\usepackage{booktabs}
\usepackage{subcaption}
\usepackage{hyperref}

\begin{document}
\mainmatter 

\title{Magnecko: Design and Control of a Quadrupedal Magnetic Climbing Robot}

\titlerunning{Magnecko: Design and Control of a Quadrupedal Magnetic Climbing Robot} 

\author{Stefan Leuthard \and Timo Eugster \and Nicolas Faesch \and Riccardo Feingold \and Connor Flynn \and Michael Fritsche \and Nicolas Hürlimann \and Elena Morbach \and Fabian Tischhauser \and Matthias Müller \and Markus Montenegro \and Valerio Schelbert \and Jia-Ruei Chiu \and Philip Arm \and Marco Hutter}

\authorrunning{S. Leuthard et al.} 


\institute{Robotic Systems Lab, ETH Zürich , 8092 Zürich, Switzerland,\\
\email{sleuthard@ethz.ch}}

\maketitle              

\begin{abstract}
Climbing robots hold significant promise for applications such as industrial inspection and maintenance, particularly in hazardous or hard-to-reach environments. This paper describes the quadrupedal climbing robot \emph{Magnecko}, developed with the major goal of providing a research platform for legged climbing locomotion. With its 12 actuated degrees of freedom arranged in an insect-style joint configuration, \emph{Magnecko}'s high manipulability and high range of motion allow it to handle challenging environments like overcoming concave \unit[90]{\textdegree} corners. A model predictive controller enables \textit{Magnecko} to crawl on the ground, on horizontal overhangs, and on vertical walls. Thanks to the custom actuators and the electro-permanent magnets that are used for adhesion on ferrous surfaces, the system is powerful enough to carry additional payloads of at least \unit[65]{\%} of its own weight in all orientations. The \emph{Magnecko} platform serves as a foundation for climbing locomotion in complex three-dimensional environments.

\keywords{Legged Robot, Climbing Robot, Inspection Robot, Magnetic Adhesion} 
\end{abstract}

\section{Introduction}

\begin{figure}[h!]
    \centering
    \begin{minipage}[b]{0.42\textwidth}
       \centering
        \includegraphics[width=.83\linewidth]{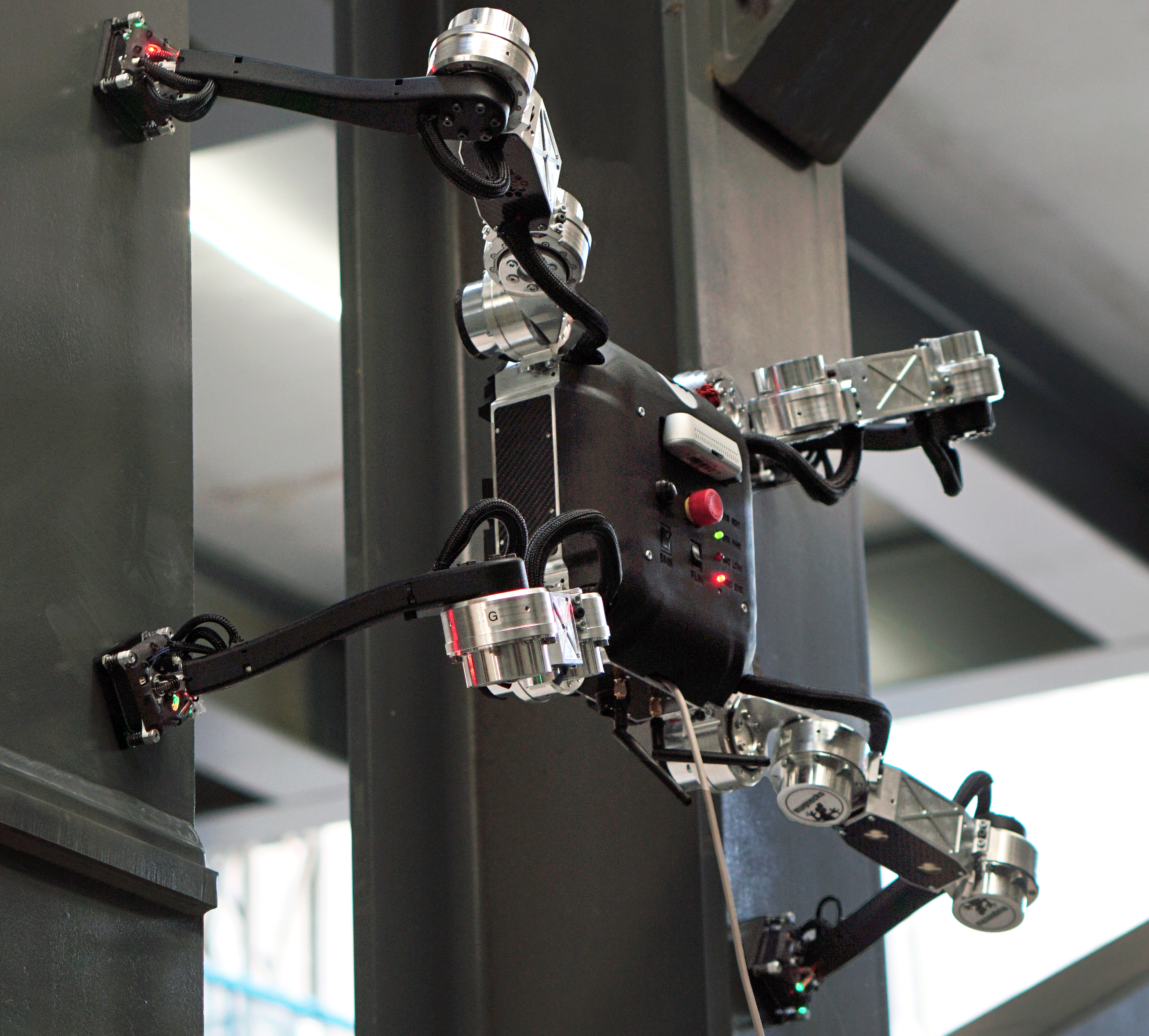}
        \caption{The magnetic climbing robot \emph{Magnecko} on two I-beams.}
        \label{fig:introduction:magnecko}
    \end{minipage}
    \hfill
    \begin{minipage}[b]{0.55\textwidth}
        \centering
        \includegraphics[width=.83\linewidth]{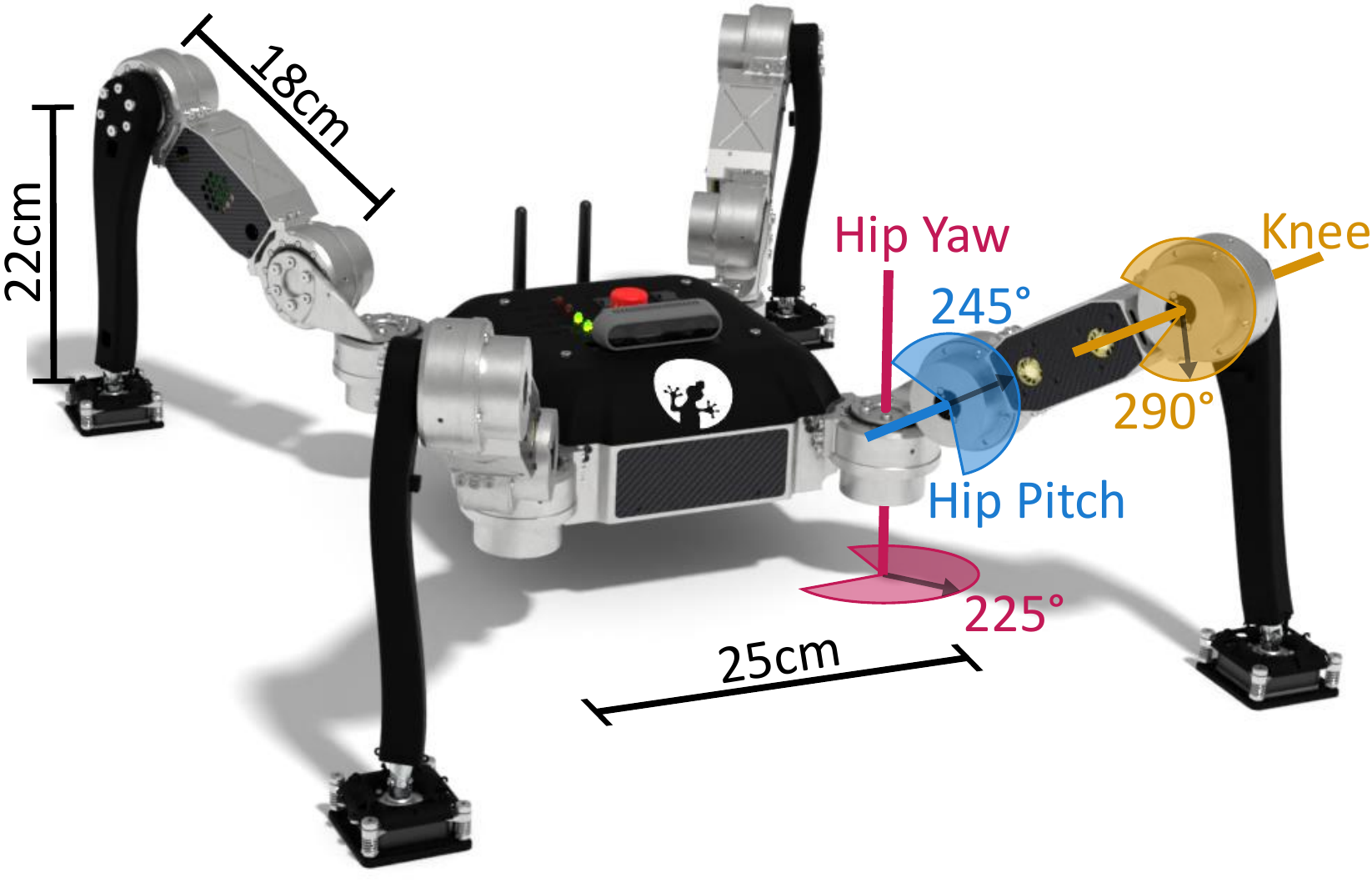}
        \caption{Insect style joint configuration of \emph{Magnecko} with the RoM and dimensions.}
        \label{fig:system:topology}
    \end{minipage}
\end{figure}

Various industrial structures, such as ship hulls, storage tanks, or bridges, demand regular inspection to identify potential safety issues such as cracks, corrosion, or welding defects. These inspections are typically labor-intensive tasks carried out by human workers and may pose substantial hazards. A climbing robot can significantly reduce this risk \cite{jose2018survey}. However, climbing robots commonly face limitations when it comes to navigating complex structures. Wheeled climbing robots like the OmniClimber \cite{Tavakoli2012} or the work by Li et al. \cite{Li2017} are often confined to continuous, unobstructed surfaces. Even though various wheeled robots are capable of overcoming ridges and transitioning between surfaces of different inclinations \cite{Fischer2008, tache2007, wang2014magnetic}, legged robots offer much more versatility to overcome a wide variety of obstacles. \\
Existing legged climbing robots use various methods of adhesion tailored to their respective operational environments. Suction cups \cite{Kang2003, hirose1991machine} and gecko-inspired adhesive grippers \cite{Parness2017} have been used to navigate smooth surfaces. To locomote on rough surfaces like rocks, spine grippers are often chosen as the adhesion method \cite{Parness2017, Uno2021}. Another commonly used option is magnetic adhesion to climb on ferrous surfaces since many industrial structures are made of steel. Existing robots have used several approaches, such as permanent magnets \cite{kamagaluh2012design}, electro-magnets \cite{grieco1998six}, and electro-permanent magnets (EPM) \cite{hong2022agile}. In this work, we decided on magnetic adhesion since this method provides a fast, strong, and reliable connection to the mating surface in controlled environments, which is ideal for a research platform. We use EPMs since they only consume power while switching the magnetic state, and they remain attached even throughout power loss.\\
Existing magnetic quadrupeds like MARVEL \cite{hong2022agile} and Magneto \cite{bandyopadhyay2018magneto} have already demonstrated the capability to climb on ferrous surfaces of various orientations. Magneto uses an admittance controller that ensures reliable de- and attachment of the magnets. The controller is however limited to relatively slow and static locomotion. In contrast, MARVEL demonstrated dynamic locomotion at high speeds using a model predictive controller (MPC) to optimize the contact forces of the stance legs and a separate PD controller to track a heuristic swing trajectory. In this work, we use an MPC approach, which additionally includes the kinematics in the formulation to control the motion of the whole body. While MARVEL used a mammal-inspired joint configuration which is advantageous for high speeds, it is not specifically designed to achieve complex maneuvers that require a large range of motion (RoM). In contrast, we chose a joint configuration with a high range of motion to enable complex climbing maneuvers. \\
Here, we present the quadrupedal climbing robot \emph{Magnecko} (Fig. \ref{fig:introduction:magnecko}), a platform designed for research on legged climbing robots. In this work, we show the first locomotion results to verify the system's capability to operate in various orientations. The system can carry high payloads, which enables attachments like inspection modules or additional sensor packages. \emph{Magnecko}'s joint topology offers a large RoM and high manipulability, which enables crossing concave \unit[90]{\textdegree} corners. Combined with the robot's torque-controllable actuation, we expect the system to allow future research on dynamic climbing locomotion on challenging structures.

\vspace{-.4cm}
\vspace{-.8cm}

\section{System Design}
\emph{Magnecko} is a quadrupedal robot with three active degrees of freedom (DOF) per leg, driven by custom actuators. A passive 3 DOF ankle joint connects a magnetic foot to each leg. The legs are mounted to the aluminum body that contains the electrical system and the onboard computer. The structural parts are mainly made out of lightweight aluminum alloys to keep the system mass low (\unit[11.3]{kg} including battery); the dimensions are shown in Fig. \ref{fig:system:topology}.

\subsection{Leg Topology}
\emph{Magnecko}'s leg topology is based on an insect-inspired joint configuration (see Fig. \ref{fig:system:topology}), as opposed to the mammalian configuration which is most commonly seen with quadrupedal robots, such as MIT Cheetah 3 \cite{bledt2018cheetah}, Spot \cite{spot} or ANYmal \cite{Hutter2017}. We chose an insect configuration because it is more efficient in climbing applications and offers higher stability since it allows lowering the body closer to the surface \cite{uno2022simulation}. Moreover, the insect configuration allows adjusting the base pose without compromising the manipulability \cite{yoshikawa1985manipulability} of the end-effectors and should, therefore, allow for more robustness while climbing in uneven terrain \cite{uno2022simulation}. \\
The large RoM, coupled with the greater manipulability of the insect topology, allows the \emph{Magnecko} platform to perform complex maneuvers in three-dimensional environments. Additionally, the symmetrical design with four identical legs eliminates a preferred locomotion direction.

\subsection{Actuator Design} 

\begin{table}[h]
    \vspace{-.8cm}
    \begin{minipage}[c]{.48\textwidth}
        \centering
        \includegraphics[width=0.95\linewidth]{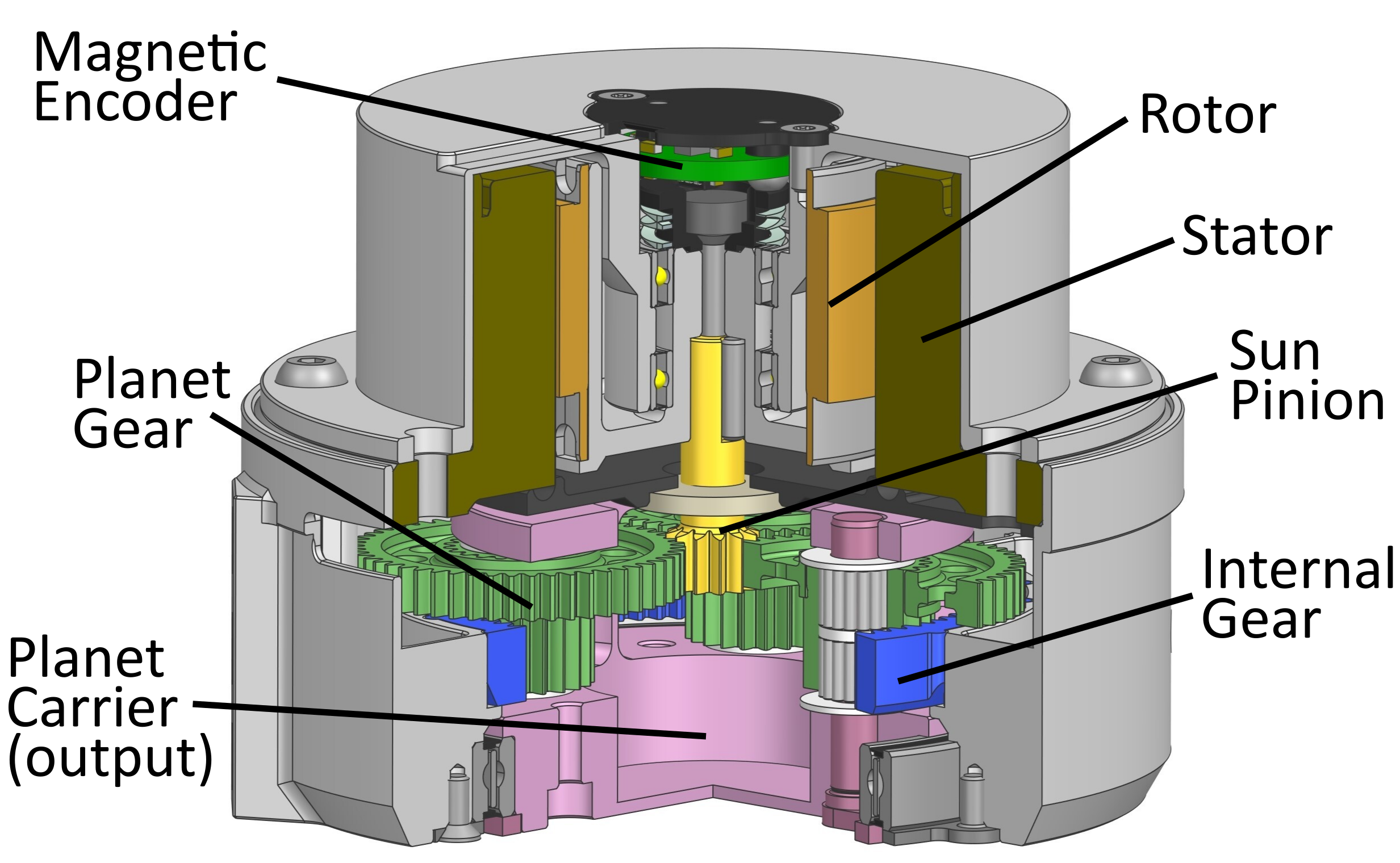}
    \end{minipage}%
    \hfill
    \begin{minipage}[c]{.48\textwidth}
        \centering
        \renewcommand{\arraystretch}{1.0}
        \begin{tabular}{@{}lcc@{}}
            \toprule
            \textbf{Parameter} \hspace{5mm} & \textbf{Value}  \\
            \midrule
             Peak Torque & $\unit[40]{Nm}$ \\
             Max Speed ($\unit[24]{V}$)& $\unit[100]{rpm}$ \\
             Mass\footnotesize{$^1$} & $\unit[510]{g}$\\
             Outer Diameter\hspace*{1.5cm}& $\unit[74]{mm}$ \\
             Axial Length & $\unit[55]{mm}$ \\
             \bottomrule
        \end{tabular}\\
        \vspace{0.1cm}\hspace*{-.6cm}\footnotesize{$^1\,$Does not include motor controller.}
        \renewcommand{\arraystretch}{1}
    \end{minipage}%
    \par
    \begin{minipage}[t]{.48\textwidth}
        \captionof{figure}{Actuator cross-section, visualizing motor, gearbox, and encoder.}
        \label{fig:system:actuator}
    \end{minipage}
    \hfill
    \begin{minipage}[t]{.48\textwidth}
        \captionof{table}{Specifications of \emph{Magnecko}'s actuators.}
        \label{tab:system:actuator}
    \end{minipage}%
    \vspace{-.4cm}
\end{table}

\noindent Each leg consists of three identical serially linked actuators. The actuators are placed directly on the joints to avoid further motion transmission that introduces undesired backlash into the system. 
Many existing legged robots use either high transmission ratio series elastic actuators \cite{Hutter2017} or low transmission ratio (up to 10:1) transparent actuators \cite{bledt2018cheetah}. In contrast, \emph{Magnecko}'s actuators consist of a custom single-stage compound planetary gearbox with an 18:1 transmission ratio, driven by a high torque density \emph{maxon DT 50 M} motor \cite{dt50m} (see Fig. \ref{fig:system:actuator}). The gearbox provides a trade-off between the robot's efficiency and high speed for dynamic locomotion while still being transparent enough to estimate the joint torques from the motor's current measurement. This estimation enables proprioception of foot reaction forces without external force or torque sensors. Table \ref{tab:system:actuator} lists the performance metrics of \emph{Magnecko}'s actuators.

\subsection{Magnetic Foot}
\begin{figure}[h]
    \vspace{-.8cm}
    \centering
    \begin{minipage}[b]{0.54\textwidth}
        \centering
        \includegraphics[width=.9\linewidth]{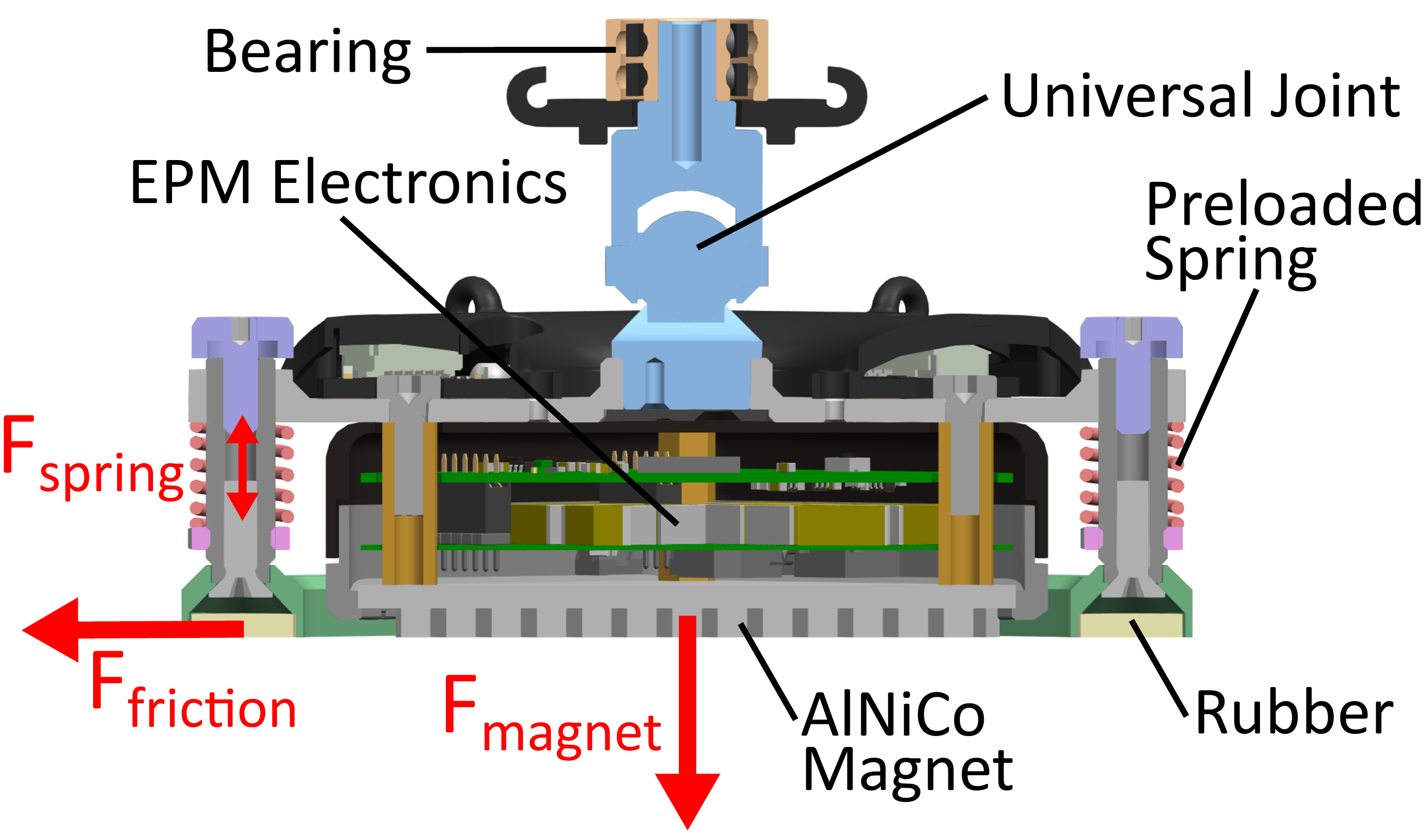}
        \caption{Section view of the magnetic foot showing the structure, ankle joint, and spring mechanism.}
        \label{fig:system:foot}
    \end{minipage}
    \hfill
    \begin{minipage}[b]{0.42\textwidth}
        \centering
        \includegraphics[width=.72\linewidth]{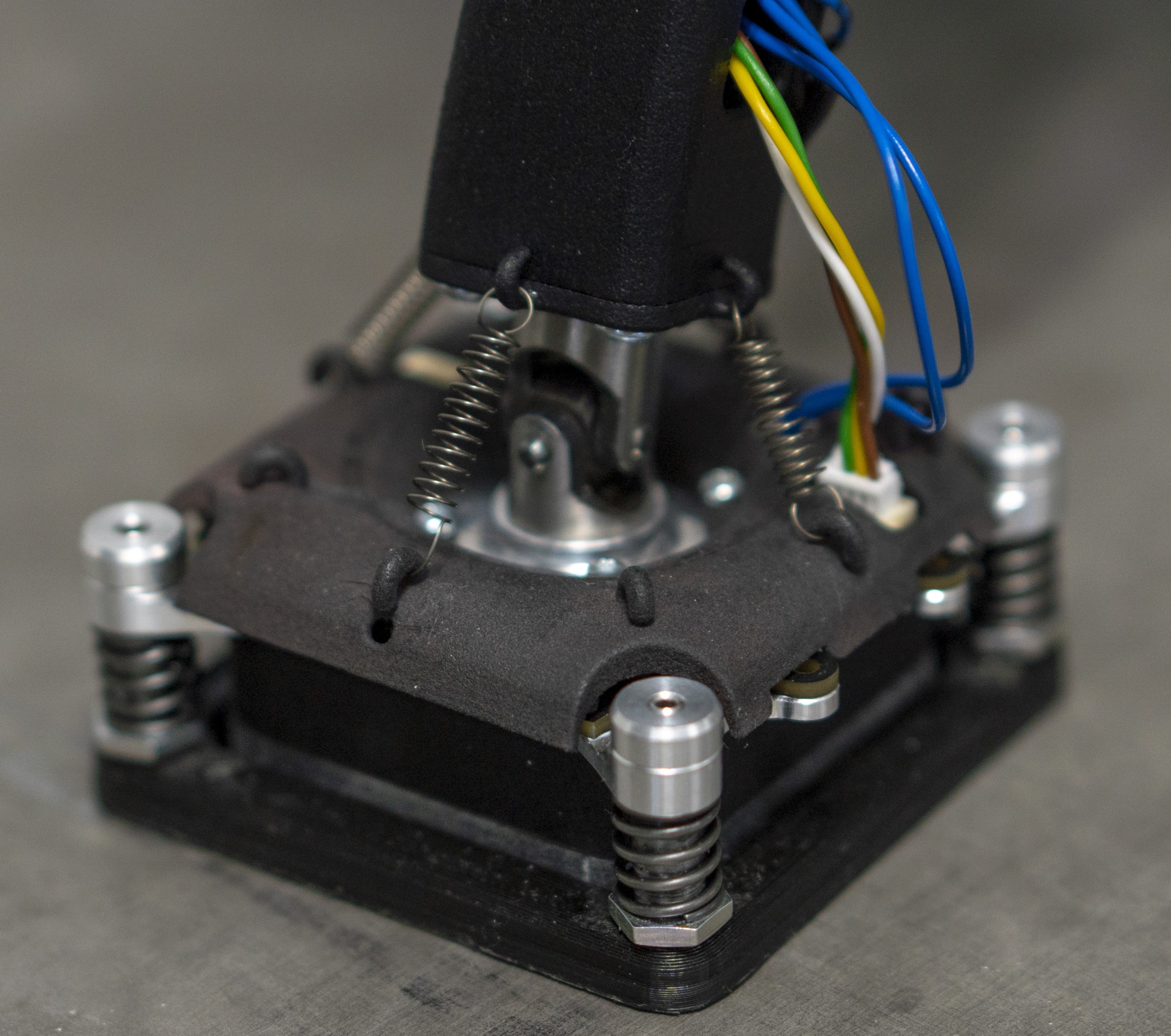}
        \caption{The fully assembled foot with springs for centering the ankle joint position.}
        \label{fig:system:foot_real}
    \end{minipage}
    \vspace{-.4cm}
\end{figure}

\noindent Each foot uses an EPM to adhere to ferrous surfaces. Since EPMs keep their magnetic state without consuming power, they are energy efficient and reduce the robot's risk of failure. We use the \emph{OpenGrab EPM v3} \cite{opengrab} with custom electronics, including a new flyback converter and higher capacitance, to increase the discharge current and decrease the switching time. We also adapted the pulsing pattern of the magnet to reach a higher level of magnetization faster. With these modifications, the EPM reaches a switching time of \unit[0.18]{s} (turn on) and \unit[0.35]{s} (turn off). However, to reach the full adhesion force, the magnet has to be magnetized twice while it is placed on the surface.\\
The EPM reaches a maximum adhesion force of \unit[380]{N} on an even surface (S235JR steel). To increase friction and, therefore, the lateral holding force, the magnet presses a rubber ring against the surface using a set of springs (see Fig. \ref{fig:system:foot}). We preloaded the springs to achieve a lateral holding force of \unit[160]{N}, which decreases the normal holding force to \unit[280]{N}. A passive 3 DOF ankle joint connects the foot to the leg. This joint allows the magnet to attach to the surface without restricting the leg from rotating around the contact point. The joint consists of a universal joint and an angular contact ball bearing. The foot can tilt up to \unit[$\pm$45]{\textdegree} around the universal joint and rotate up to \unit[$\pm$90]{\textdegree} around the bearing. Springs between the leg and the foot ensure that the foot returns to a defined neutral position when the leg is in motion (see Fig. \ref{fig:system:foot_real}).

\subsection{Electrical System} 
The system runs at a voltage of \unit[24]{V} which can be provided either through a cable or by an onboard 6-cell lithium-polymer battery that allows a runtime of \unit[45]{minutes}. A power distribution board supplies the actuators with power and includes DC/DC converters to provide additional voltage levels. To run the controls onboard the robot, we integrated an Intel NUC11TNBv7 with an Intel Core i7-1185G7 processor. The body also houses the Vectornav VN100 \cite{vectorNAV} Inertial Measurement Unit (IMU). A front-facing depth camera (Intel Realsense D435 \cite{real}) is mounted on top of the body. To control our actuators, we use the ezMotion MMP545400-75-E2-1 \cite{mc} motor controllers. The motor controllers of the four hip yaw actuators are placed inside the body. Each upper leg contains two motor controllers for the hip pitch and knee actuators. The motor controllers allow us to estimate the joint torque from the measured phase currents.

\section{Modelling and Control}

\begin{figure}[h]
    \vspace{-0.8cm}
    \centering
    \includegraphics[width=0.95\linewidth]{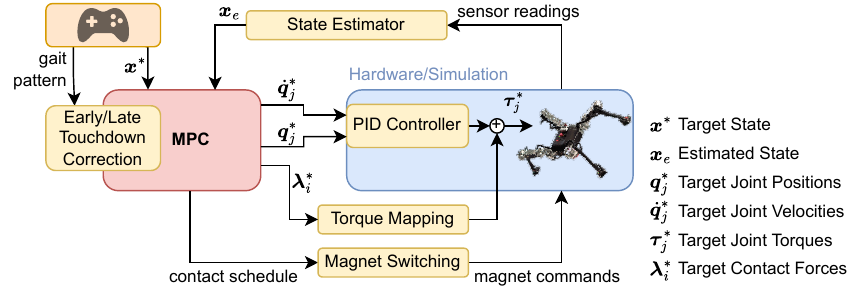}
    \caption{Controller architecture running onboard the robot. The operator commands are sent from a separate computer over a wireless network.}
    \label{fig:control:architecture}
    \vspace{-.5cm}
\end{figure}

\noindent The onboard control framework (see Fig. \ref{fig:control:architecture}) uses ROS2 \cite{ros} and is implemented in C++. We use ros2\_control \cite{ros2_control} to abstract the communication with the hardware which allows us to switch between the simulation and the physical hardware seamlessly. An operator commands a target state and selects a predefined gait pattern. Based on this pattern the controller extracts a contact schedule and updates it continuously to correct for early or late touchdowns. A model predictive controller (MPC) calculates an optimized motion plan at a frequency of approximately \unit[130]{Hz} to track the operator's command. We initialize each MPC iteration with the estimated state from the open source state estimator pronto \cite{camurri2020pronto} at a frequency of \unit[400]{Hz}. Our motor controllers finally track the optimized motion plan using a PID controller on the target joint position and velocity. Additionally, the controller maps the optimized ground reaction forces (GRFs) $\boldsymbol{\lambda}_i$ to joint torques and applies them as a feed-forward term. The controller sends the commands to switch the magnet state at a fixed offset to the planned liftoff/touchdown times to account for the magnet switching time.

\subsection{Model Predictive Control}

We used the ocs2 toolbox \cite{ocs2} to implement the MPC. The formulation of the MPC problem is mainly based on the work by Grandia et al. \cite{grandia2023perceptive}. The controller solves the optimization problem

\begin{equation}
\begin{aligned}
\min_{\boldsymbol{u}(\cdot)} \qquad &  \int_{t_0}^{t_0 + T} L(\boldsymbol{x}(t), \boldsymbol{u}(t), t) \,dt , \\
\textrm{s.t.} \qquad &    \boldsymbol{x}(t_0) = \boldsymbol{x}_e , \\
    &                    \dot{\boldsymbol{x}} = \boldsymbol{f}(\boldsymbol{x}(t), \boldsymbol{u}(t), t) ,    \\
    &                    \boldsymbol{g}(\boldsymbol{x}(t), \boldsymbol{u}(t), t) = \boldsymbol{0} ,   \\
\end{aligned}
\end{equation}

\noindent where $\boldsymbol{x}(t)$ and $\boldsymbol{u}(t)$ are the state and input at time $t$. $T$ is the time horizon of the MPC and $\boldsymbol{x_e}$ is the latest estimated state. The aim is to find an input $\boldsymbol{u}(t) = \begin{bmatrix} \boldsymbol{\lambda}_1^T & \hdots & \boldsymbol{\lambda}_4^T & \dot{\boldsymbol{q}}_j^T\end{bmatrix}^T$ that minimizes the cost-term while fulfilling the system dynamics $\boldsymbol{f}(\boldsymbol{x}, \boldsymbol{u}, t)$ and equality constraints $\boldsymbol{g}(\boldsymbol{x}, \boldsymbol{u}, t)$. We include all inequality constraints in the cost term using a relaxed barrier penalty function \cite{hauser2006barrier}. To solve the optimization problem, we use the sequential quadratic programming solver described in the work by Grandia et al. \cite{grandia2023perceptive}. 

\noindent We model the system dynamics using simplified single rigid body dynamics (SRBD). We assume a constant center of mass (CoM) and a constant inertia tensor; both are evaluated only once for a default leg configuration. Since the robot needs to operate in various orientations, we formulate the rotational dynamics using the singularity-free unit quaternion $\boldsymbol{\xi}$ \cite{graf2008quaternions} instead of Euler angles. To obtain the swing trajectories directly from the MPC solution, we model the kinematics of the system by including the joint positions $\boldsymbol{q}_j$ in the state vector. We can, therefore, formulate the dynamics as

\begin{equation}
    \boldsymbol{f}(\boldsymbol{x},\boldsymbol{u}, t) = \frac{d}{dt}
    \begin{bmatrix}
        \boldsymbol{p} \\
        \boldsymbol{v} \\
        \boldsymbol{\xi} \\
        \boldsymbol{\omega} \\
        \boldsymbol{q}_j
    \end{bmatrix}
    =
    \begin{bmatrix}
        \boldsymbol{v} \\
        \boldsymbol{g} + \frac{1}{m} \sum\limits_{i=1}^{4} \boldsymbol{\lambda}_i \\
        \frac{1}{2} \boldsymbol{\xi} \circ \boldsymbol{\xi}_{\omega} \\
        \boldsymbol{I}^{-1} (\boldsymbol{C}(\boldsymbol{\xi})^T( \sum\limits_{i=1}^{4} \boldsymbol{r}_i \times \boldsymbol{\lambda}_i) - \boldsymbol{\omega} \times (\boldsymbol{I}\boldsymbol{\omega})) \\
        \dot{\boldsymbol{q}}_j
    \end{bmatrix},
\end{equation}

\noindent where $\boldsymbol{p}$ and $\boldsymbol{v}$ are the position and linear velocity of the base and $\boldsymbol{\omega}$ is the angular velocity. $\boldsymbol{I}$ is the inertia tensor, $\boldsymbol{C}(\boldsymbol{\xi})$ is the rotation matrix representation of the base orientation and $\boldsymbol{r}_i$ is the vector from the center of mass to the $i$-th end-effector. The operator $\circ$ represents the quaternion product, where $\boldsymbol{\xi}_{\omega} = (0, \boldsymbol{\omega})$.

\noindent The quadratic tracking cost

\begin{equation}
    L_t = \frac{1}{2} \boldsymbol{\epsilon}_x^T \boldsymbol{Q} \boldsymbol{\epsilon}_x + \frac{1}{2} \boldsymbol{\epsilon}_u^T \boldsymbol{R} \boldsymbol{\epsilon}_u,
\end{equation}
penalizes the deviations $\boldsymbol{\epsilon}_x$, $\boldsymbol{\epsilon}_u$ from the target state and input, where $\boldsymbol{Q}$ and $\boldsymbol{R}$ are weighting matrices. The controller determines the target state based on the operator commands, and we choose the target input to distribute the total mass equally over all stance legs with zero joint velocities. To calculate the deviation from the target orientation, we use the quaternion error as defined by Siciliano et al. \cite{Siciliano2009}. 

\noindent To enforce the contact schedule, we constrain the velocity at all stance end-effectors to be zero and enforce swing trajectories in the normal direction of the surface. We also define a set of constraints on the GRFs. While the contact force of the swing legs must be zero, the forces of the stance legs need to fulfill the friction cone constraint

\begin{equation}
    \mu(F_{z,i} + F_{mag}) - \sqrt{F_{x,i}^2 + F_{y,i}^2 + \epsilon} \geq 0 \text{,} \qquad \text{with } \begin{bmatrix} F_{x,i} & F_{y,i} & F_{z,i}\end{bmatrix}^T = \boldsymbol{C}_{TI} \cdot \boldsymbol{\lambda}_i
\end{equation}
to avoid slipping. We rotate the cone according to the surface inclination by formulating the constraint in the local terrain frame, where $\boldsymbol{C}_{TI}$ is the rotation matrix from the inertial to the terrain frame. $F_{mag} > 0$ is the magnetic adhesion force, $\mu$ is the friction coefficient and $\epsilon > 0$ ensures a finite derivative. To guarantee that the magnet reliably attaches to the surface, we impose the additional constraint $F_{z,i} - F_{min} \geq 0$. The constraint is only active for a short period after the touchdown and ensures the foot is pressed against the surface with a force of at least $F_{min} = \unit[20]{N}$ during the magnetization of the EPM. Finally, we constrain upper and lower bounds on the joint positions, velocities, and torques.

\subsection{Transition Control}
To showcase the platform's capability to locomote in more complex environments than on flat surfaces, we implement a simple controller to transition between \unit[90]{\textdegree} concave corners. To detect upcoming surfaces, we use the point cloud from the front-facing depth camera and fit a plane to the central patch of points. The controller then generates a heuristic motion plan by calculating polynomial trajectories to the target footholds. We select these footholds such that the distance between the front and hind legs matches a defined value. A differential inverse kinematics (IK) controller finally tracks the motion plan.

\section{Results}
We conducted our experiments on flat steel surfaces, testing locomotion on the ground, ceiling, and vertical walls. The controller solves the optimization problem for a time horizon of $T = \unit[1.0]{s}$ with a nominal time discretization of $\delta t = \unit[0.02]{s}$.
These values ensure the horizon covers at least one swing duration and the controller operates at approximately \unit[130]{Hz}. In addition to the MPC, we used a differential inverse kinematics (IK) controller for comparison.

\subsection{Locomotion Results}

\begin{figure}[h]
     \centering
         \includegraphics[width=\textwidth]{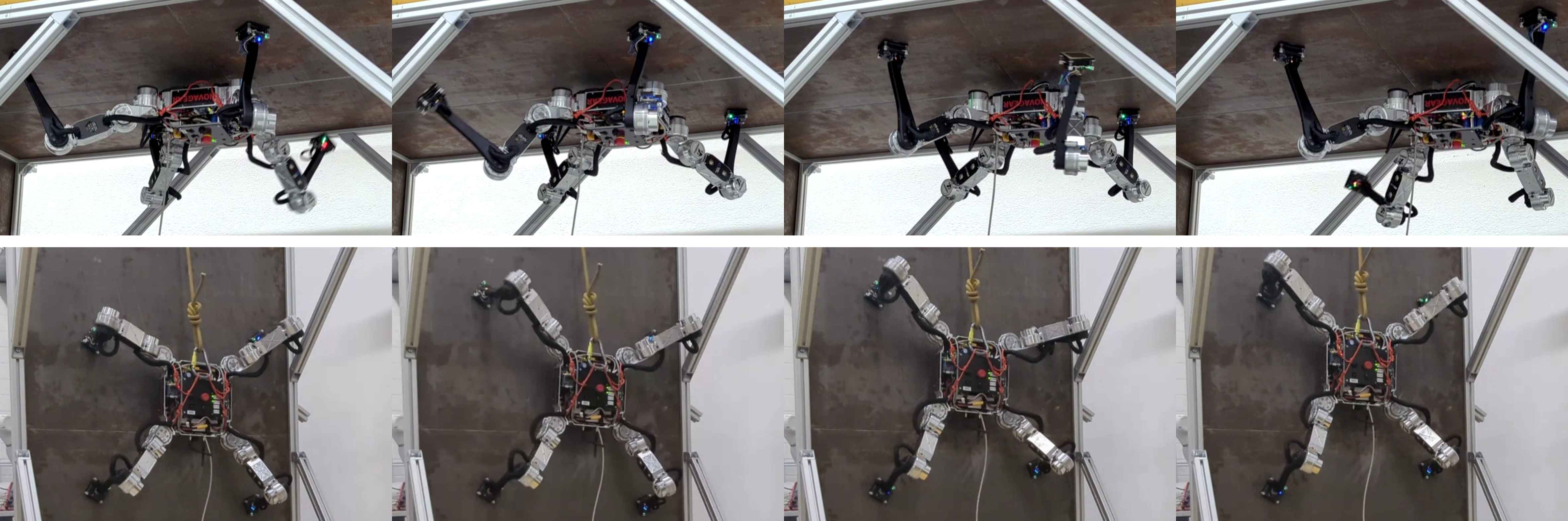}
         \caption{Image sequence of the climbing experiments on the ceiling and wall. Videos of our experiments are provided at \url{https://youtu.be/hL1TrDVoaa8}.}
         \label{fig:results:crawling}
         \vspace{-.4cm}
\end{figure}

We successfully demonstrated \emph{Magnecko}'s capability of crawling on the ground, on vertical walls, and on the ceiling (see Fig. \ref{fig:results:crawling}). In between two steps, the robot magnetizes and demagnetizes the magnets successively, so three feet are always attached with the full adhesion force. As a result, all four feet are in contact for a period of \unit[0.6]{s} before the next swing leg lifts off. On the ground, the robot does not require the full adhesion force, and a full contact period of \unit[0.2]{s} is sufficient. The robot, therefore, reaches the maximum speed of \unit[0.15]{m/s} on the ground. Compared to the baseline IK controller, the MPC increases the maximum speed by a factor of 2 up to a factor of 4.5 (see Table \ref{tab:results:crawling:speed:mpc}). \\
To verify the choice of the EPMs, we evaluated the margin of the GRFs to a detachment. We observed a maximum normal pulling force of \unit[100]{N} when climbing on the ceiling, which is only $\sim$\unit[35]{\%} of the maximum of \unit[280]{N}. If tangential forces are present, the foot will fail by slipping before it tilts. The experiments showed that the robot operates with a large margin to slipping (see Fig. \ref{fig:results:grf}). However, we found that the adhesion force drastically decreases when the surface is not perfectly flat (e.g. cracks, dirt, or corrosion).\\

\begin{table}[h]
    \vspace{-1cm}
    \begin{minipage}[c]{.48\textwidth}
        \centering
        \renewcommand{\arraystretch}{1.5}
        \begin{tabular}{@{}lcc@{}}
            \toprule
            \textbf{Orientation} & \textbf{MPC} & \textbf{IK}  \\
            \midrule
             Ground & $\unit[0.15]{\frac{m}{s}}$ & $\unit[0.033]{\frac{m}{s}}$ \\
             Ceiling & $\unit[0.058]{\frac{m}{s}}$ & $\unit[0.016]{\frac{m}{s}}$ \\
             Wall & $\unit[0.033]{\frac{m}{s}}$ & $\unit[0.016]{\frac{m}{s}}$ \\
             \bottomrule
        \end{tabular}
        \renewcommand{\arraystretch}{1}
    \end{minipage}%
    \hfill
    \begin{minipage}[c]{.48\textwidth}
        \centering
        \includegraphics[width=1.0\linewidth]{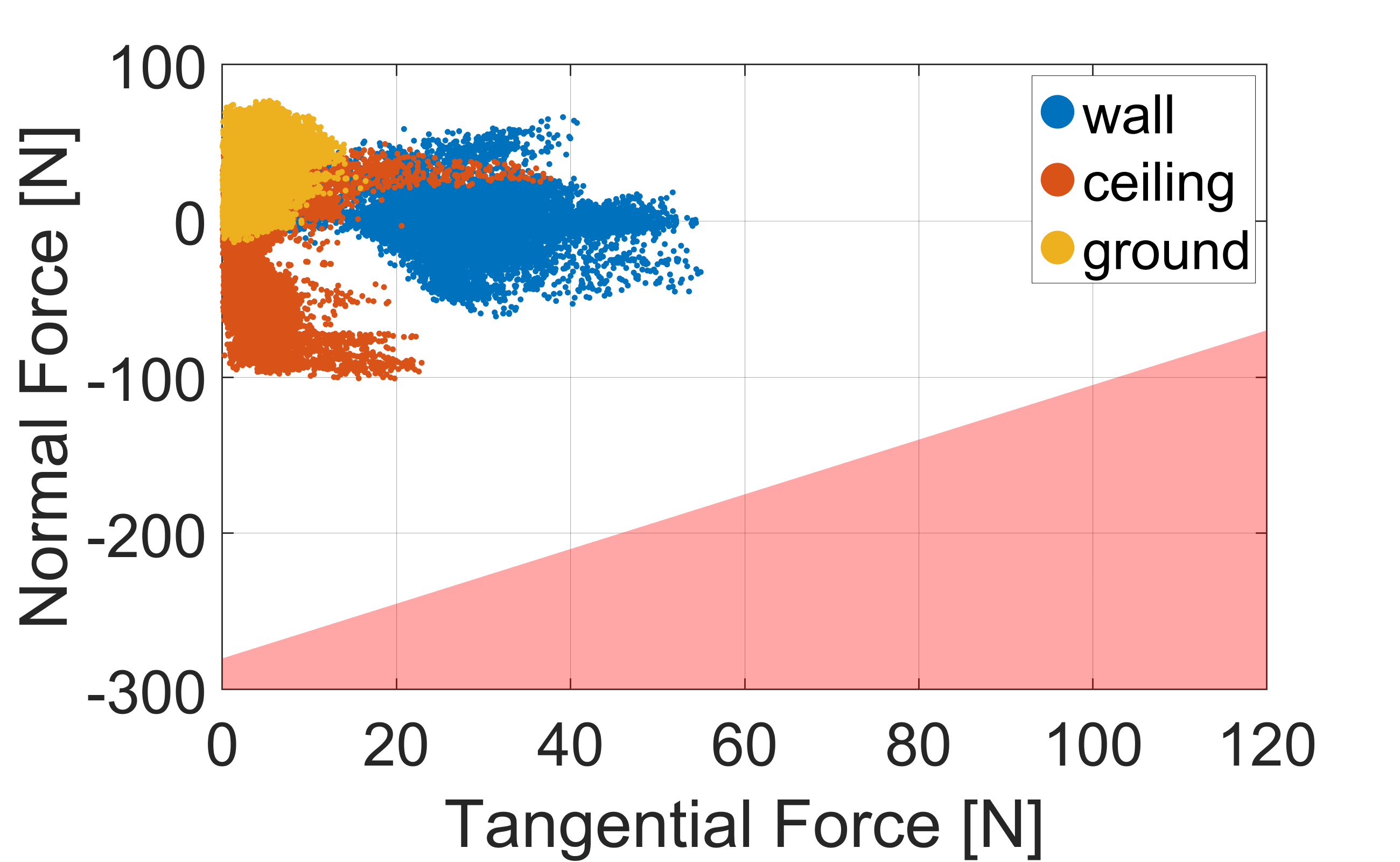}
    \end{minipage}%
    \par
    \begin{minipage}[t]{.48\textwidth}
        \caption[Maximum Speed Using a Crawling Gait]{Maximum speeds achieved at different inclinations using the MPC and the baseline IK controller. 
        }
        \label{tab:results:crawling:speed:mpc}
    \end{minipage}
    \hfill
    \begin{minipage}[t]{.48\textwidth}
        \captionof{figure}{Normal and tangential contact forces estimated from the observed joint torques. The red area marks forces outside the friction cone.}
        \label{fig:results:grf}
    \end{minipage}%
    \vspace{-.5cm}
\end{table}

\begin{figure}[h!]
    \vspace{-.5cm}
    \centering
    \includegraphics[width=1.0\linewidth]{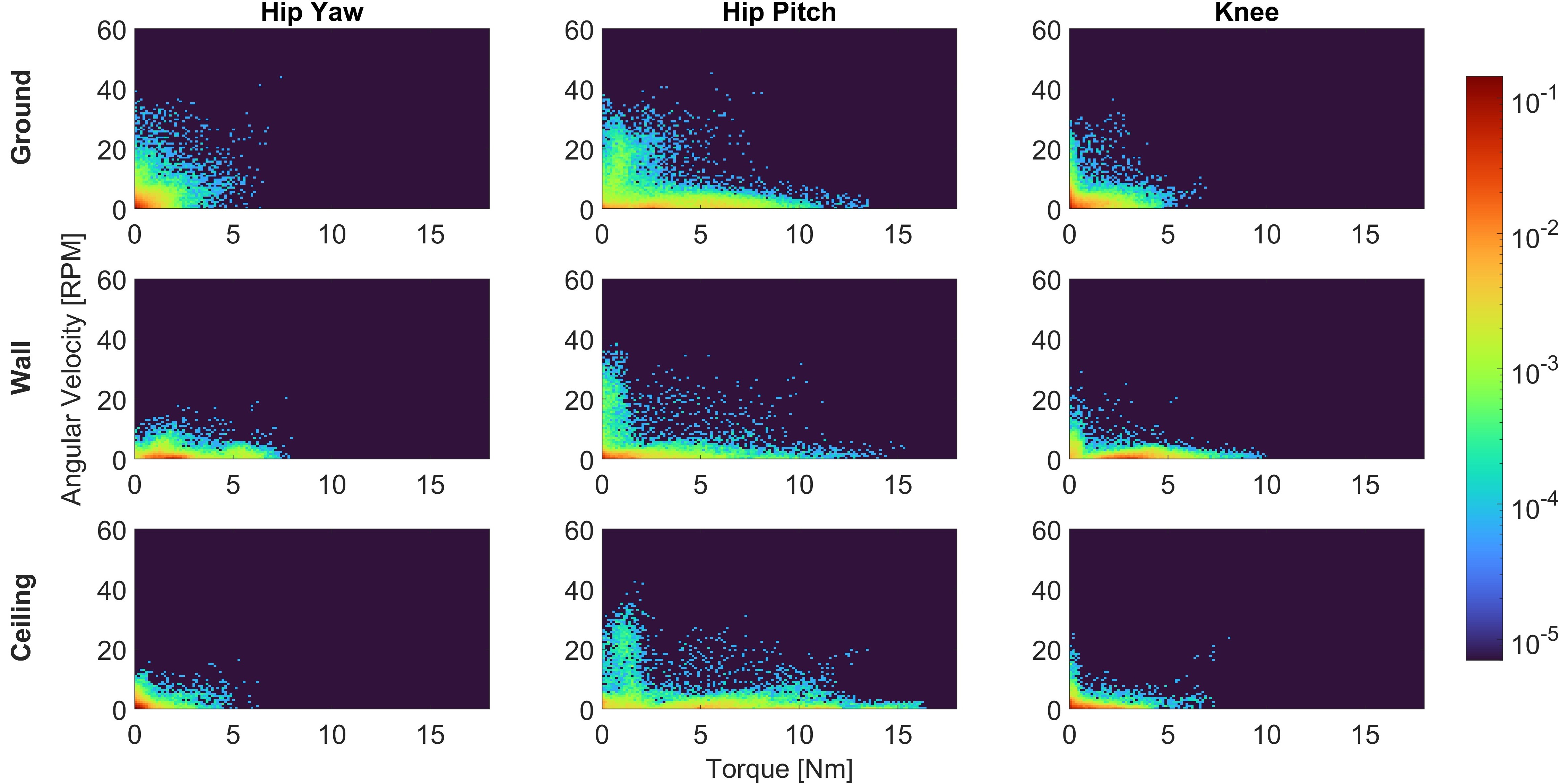}
    \caption{Probability distribution of joint torques and velocities during locomotion in the three orientations.}
    \label{fig:results:TorquesVelocities}
    \vspace{-.7cm}
\end{figure}

\begin{figure}[h]
    \vspace{-.7cm}
    \centering
    \includegraphics[width=0.85\linewidth]{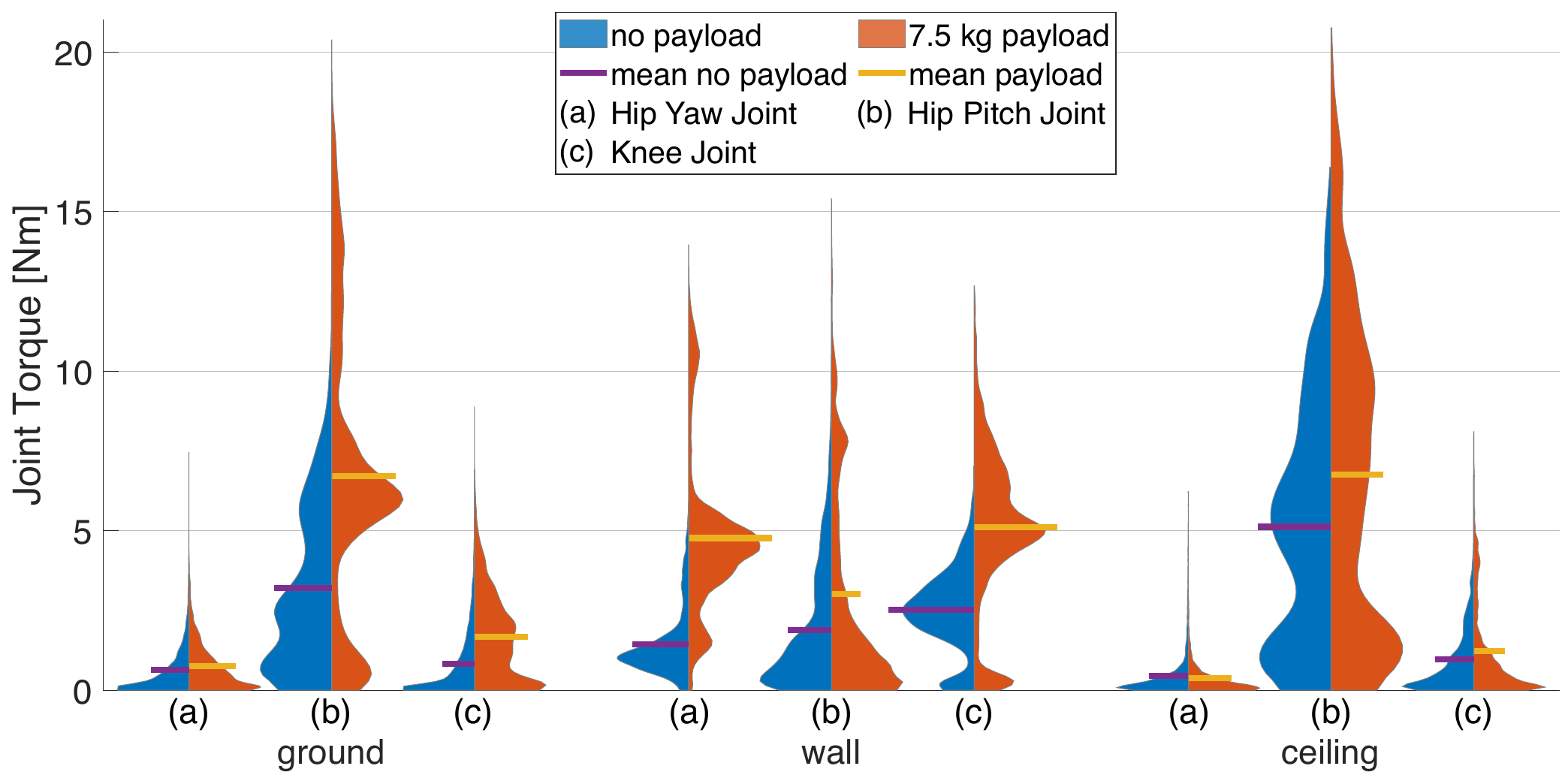}
    \caption{Torque distribution with \unit[7.5]{kg} payload compared to no palyoad.}
    \label{fig:results:payload}
    \vspace{-.7cm}
\end{figure}

\noindent In our experiments, the joint torques and velocities constantly remained far from their limits (see Fig. \ref{fig:results:TorquesVelocities}). We observed a maximum torque of $\sim$\unit[16]{Nm}, while our actuators could theoretically deliver torques of up to \unit[40]{Nm}. The hip pitch actuators experience the largest load. On the ceiling, they exert \unit[5.1]{Nm} on average, which is well below the maximum continuous torque since the motor winding temperatures remain stable and below \unit[40]{°C} at ambient room temperature. Additionally, we observed actuator speeds of up to \unit[40]{rpm}, while the actuators can achieve up to \unit[100]{rpm}. \\
Since \emph{Magnecko} operates far from the limits of the actuators and the magnetic feet, the robot can carry additional payloads. We successfully tested payloads up to \unit[7.5]{kg} on the ground, wall, and ceiling. Even with the additional payload, the actuators operate far from their limits (see Fig. \ref{fig:results:payload}). 

\subsection{Transition Between Different Inclinations}

\begin{figure}[h]
    \centering
    \includegraphics[width=1.0\textwidth, trim={0 105.8mm 0 0}, clip]{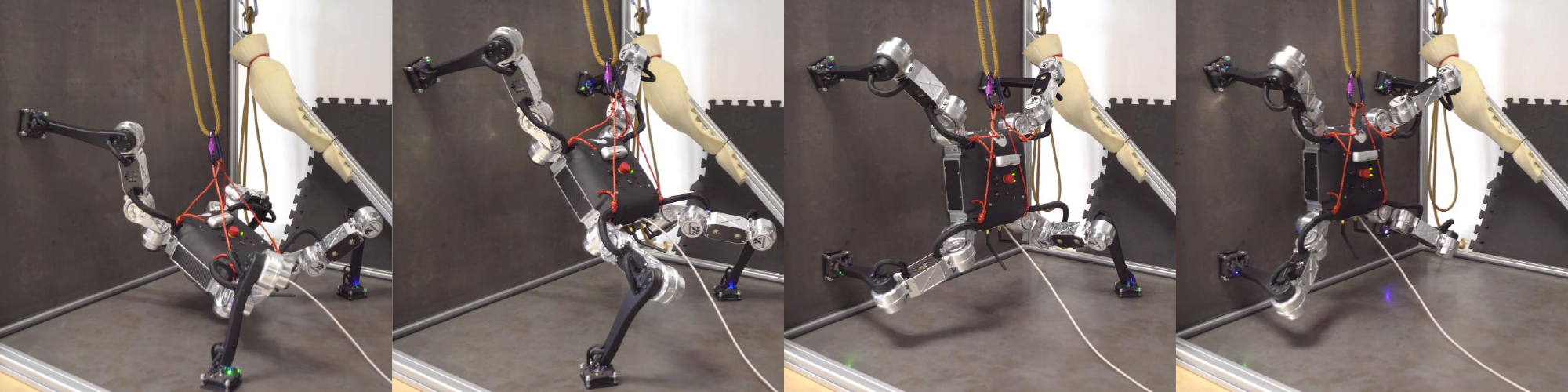}
    \caption{\emph{Magnecko} performing a transition from the ground to the wall.}
    \label{fig:results:TransitionImage}
    \vspace{-.4cm}
\end{figure}

\noindent We successfully demonstrated \emph{Magnecko}'s capability to transition from ground to wall, wall to ceiling, ceiling to wall, and wall to ground (see Fig. \ref{fig:results:TransitionImage} and supplementary video at \url{https://youtu.be/hL1TrDVoaa8}). The large RoM of the insect-style joint configuration allows \emph{Magnecko} to transition with only six steps. In our experiments, a single transition took approximately \unit[18]{s}. 

\section{Discussion}

The crawling gait does not exhaust the capabilities of our actuators. The large margins of the joint torques and velocities will allow us to research faster motions and dynamic gaits. Additionally, the resulting payload capabilities enable us to explore different attachments, like modules for inspection or maintenance work. Currently, the limitation regarding payload is not the hardware but the controller, since we did not model the additional weight. Consequently, we observed large tracking errors in the base pose with \unit[7.5]{kg} payload attached.

\noindent While our EPMs can support the increased GRFs when carrying a payload, they are limited to completely flat and clean surfaces. To deploy \emph{Magnecko} in real-world environments, it will be essential that the magnetic feet are robust to imperfections in the surface. Additionally, the switching durations of our EPMs are the main speed limitation. A faster switching time would allow a shorter full stance phase and increase \emph{Magnecko}'s speed. Another limitation regarding speed is the maximum step size. While the RoM would allow for much larger step sizes, the foot touches down with a tilt angle that does not guarantee alignment with the surface if the step size is too large. This issue is most pronounced on the wall since here, gravity deflects the foot orientation the most. Consequently, we reached the lowest speed in this orientation.

\noindent While we specifically chose \emph{Magnecko}'s leg configuration to enable locomotion in complex environments the current controller and sensor package limits the locomotion to primitive environments that only consist of flat surfaces and \unit[90]{\textdegree} concave corners. To navigate more complex real-world environments such as beams, confined spaces, or gaps, the single front-facing depth camera will not suffice. Additional perceptive sensors and a perception pipeline will be necessary to allow \emph{Magnecko} to extract suitable foothold locations and enable localization and path planning. 

\section{Conclusion and Future Work}
In conclusion, this work introduces the quadrupedal climbing robot \emph{Magnecko} as a research platform for legged climbing locomotion. We successfully demonstrated the platform's capability to crawl on the ground, vertically, and on overhanging surfaces. Using the MPC, we achieved speeds of \unit[0.15]{m/s} on the ground, \unit[0.058]{m/s} on a horizontal overhang and \unit[0.033]{m/s} on vertical walls. 
Moreover, \emph{Magnecko} can overcome \unit[90]{\textdegree} concave corners. While we only demonstrated a crawling gait in this work, we specifically designed the hardware, especially our actuators, to enable us to explore dynamic gaits. 

\noindent In future work, we will explore locomotion in more complex environments. The work will include the integration of a perception pipeline to select suitable footholds. Additionally, we will integrate a more robust foot design that can adapt to the geometry of the underlying surface.

\bibliographystyle{bib/splncs03.bst}
\bibliography{bib/main}

\end{document}